\documentclass{article}

\usepackage[final,nonatbib]{nips_2018}

\usepackage[utf8]{inputenc} 
\usepackage[T1]{fontenc}    
\usepackage{hyperref}       
\usepackage{url}            
\usepackage{booktabs}       
\usepackage{amsfonts}       
\usepackage{nicefrac}       
\usepackage{microtype}      

\usepackage{amsmath}
\usepackage{amssymb}
\usepackage{multirow}
\usepackage{xcolor}

\newcommand{\bftheta}{\pmb{\theta}}

\newcommand{\bfz}{\mathbf{z}}
\newcommand{\bfphi}{\pmb{\phi}}
\newcommand{\bfpi}{\pmb{\pi}}
\newcommand{\bfalpha}{\pmb{\alpha}}
\newcommand{\EE}{\mathop{\mathbb{E}}}
\newcommand{\blue}[1]{\textcolor{blue}{#1}}
\newcommand{\bfblue}[1]{{\bf \blue{#1}}}

\title{FLOPs as a Direct Optimization Objective for Learning Sparse Neural Networks}

\author{
  Raphael Tang, Ashutosh Adhikari, Jimmy Lin\\
  David R. Cheriton School of Computer Science\\
  University of Waterloo\\
  \texttt{\{r33tang, ashutosh.adhikari, jimmylin\}@uwaterloo.ca}
}

\begin{document}

\maketitle

\begin{abstract}
There exists a plethora of techniques for inducing structured sparsity in parametric models during the optimization process, with the
final goal of resource-efficient inference. However, few methods target a specific number of floating-point operations 
(FLOPs) as part of the optimization objective, despite many
reporting FLOPs as part of the results. Furthermore, a one-size-fits-all approach ignores realistic system constraints, which differ 
significantly between, say, a GPU and a mobile phone---FLOPs on the former
incur less latency than on the latter; thus, it is important for practitioners to be able to specify a target number of FLOPs during
model compression. In this work, we extend a state-of-the-art technique to directly incorporate
FLOPs as part of the optimization objective and show that, given a desired FLOPs requirement, different neural networks can be successfully 
trained for image classification.
\end{abstract}

\section{Introduction}

Neural networks are a class of parametric models that achieve the state of the art across a broad range of tasks, but their heavy computational requirements hinder practical deployment
on resource-constrained devices, such as mobile phones, Internet-of-things (IoT) devices, and offline embedded systems. Many recent works focus on 
alleviating these computational burdens, mainly falling under two non-mutually exclusive categories: manually designing resource-efficient models,
and automatically compressing popular architectures. In the latter, increasingly sophisticated 
techniques have emerged~\cite{li2017pruning, liu2017learning, louizos2017bayesian}, which have achieved respectable accuracy--efficiency operating points, 
some even Pareto-better than that of the original network; for example, network slimming~\cite{li2017pruning} reaches an error rate of 6.20\% on CIFAR-10 using VGGNet~\cite{simonyan2014very} with a 51\% FLOPs reduction---an error decrease of 0.14\% over the original.

However, few techniques impose a FLOPs constraint as part of a single optimization objective. 
Budgeted super networks~\cite{veniat2018learning} are closely related to this work, incorporating FLOPs and memory usage objectives
as part of a policy gradient-based algorithm for learning sparse neural architectures.
MorphNets~\cite{gordon2018morphnet} apply an $L_1$ norm, shrinkage-based relaxation of a FLOPs objective, but for the purpose of
searching and training multiple models to find good network architectures; in this work, we learn a sparse neural network in a single training run.
Other papers directly target device-specific metrics, such as energy usage~\cite{yang2017designing}, but the pruning procedure does not explicitly
include the metrics of interest as part of the optimization objective, instead using them as heuristics. Falling short of continuously
deploying a model candidate and measuring actual inference time, as in time-consuming neural architectural search~\cite{tan2018mnasnet}, we believe that the number of FLOPs
is reasonable to use as a proxy measure for actual latency and energy usage; across variants of the same
architecture, Tang et al. suggest that the number of FLOPs is a stronger predictor of energy usage and latency than the number of parameters~\cite{tang2018experimental}. 

Indeed, there are compelling reasons to optimize for the number of FLOPs as part of the training objective: First, 
it would permit FLOPs-guided compression in a more principled manner.
Second, practitioners can directly specify a desired target of FLOPs, which is important in deployment. Thus, our main contribution
is to present a novel extension of the prior state of the art~\cite{louizos2018learning} to incorporate the number of FLOPs as part of
the optimization objective, furthermore allowing practitioners to set and meet a desired compression target.

\section{FLOPs Objective}

Formally, we define the FLOPs objective $L_{flops} : f \times \mathbb{R}^{m} \mapsto \mathbb{N}_0$ as follows:
\begin{equation}
L_{flops}(h, \bftheta) := g(h(\cdot; \mathbb{I}(\bftheta_1 \neq 0), \mathbb{I}(\bftheta_2 \neq 0), \dots, \mathbb{I}(\bftheta_m \neq 0))) \qquad\qquad |\bftheta| = m
\end{equation}
where $L_{flops}$ is the FLOPs associated with hypothesis $h(\cdot; \bftheta) := p(\cdot | \bftheta)$, $g(\cdot)$ is 
a function with the explicit dependencies, and $\mathbb{I}$ is the indicator function. We assume
$L_{flops}$ to depend only on whether parameters are non-zero, such as the number of neurons in a 
neural network. For a dataset $\mathcal{D}$, our empirical risk thus becomes
\begin{equation}
\mathcal{R}(h; \bftheta) = -\log{p(\mathcal{D}|\bftheta)} + \lambda_f \max\left(0, L_{flops}\left(h, \bftheta\right) - T\right) \qquad \mathcal{D} = ((x_1, y_2), \dots, (x_n, y_n))
\end{equation}
Hyperparameters $\lambda_f \in \mathbb{R}^+_0$ and $T \in \mathbb{N}_0$ control the strength of the FLOPs objective and the target, respectively. The 
second term is a black-box function, whose combinatorial nature prevents gradient-based optimization; thus, using the same procedure in prior art~\cite{louizos2018learning}, we 
relax the objective to a surrogate of the evidence lower bound with a fully-factorized spike-and-slab posterior as the variational distribution, 
where the addition of the clipped FLOPs objective can be interpreted as a sparsity-inducing prior $p(\bftheta) \propto \exp(-\lambda_f \max(0, L_{flops}(h, \bftheta) - T))$. 
Let $\bfz \sim p(\bfz | \bfpi)$ be Bernoulli random variables parameterized by $\bfpi$:
\begin{equation}
\mathcal{L}(h; \bftheta) = \EE_{p(\bfz | \bfpi)}\left[-\log{p(\mathcal{D}|\bftheta \odot \bfz)} + \lambda_f \max\left(0, L_{flops}\left(h,\bftheta \odot \bfz\right) - T\right)\right] \label{eq:lat1}
\end{equation}
where $\odot$ denotes the Hadamard product. To allow for efficient reparameterization and exact zeros, Louizos et al.~\cite{louizos2018learning} propose to use a hard concrete distribution as 
the approximation, which is a stretched and clipped version of the binary Concrete distribution~\cite{maddison2016concrete}: if 
$\hat{z} \sim \text{BinaryConcrete}(\alpha, \beta)$, then $\tilde{z} := \max(0, 
\min(1, (\zeta - \gamma) \hat{z} + \gamma))$ is said to be a hard concrete r.v., given $\zeta > 1$ and $\gamma < 0$. Define $\bfphi := (\bfalpha, \beta)$, and let
$\psi(\bfphi) = \text{Sigmoid}(\log \bfalpha - \beta \log \frac{-\gamma}{\zeta})$ and $\bfz \sim \text{Bernoulli}(\psi(\bfphi))$. Then, the approximation becomes
\begin{equation}
\mathcal{L}(h; \bftheta) \approx \EE_{p(\tilde{\bfz} | \bfphi)}\left[-\log{p(\mathcal{D}|\bftheta \odot \bfz)}\right] + \lambda_f \EE_{p(\bfz | \psi(\bfphi))}\left[\max\left(0, L_{flops}\left(h,\bftheta \odot \bfz\right) - T\right)\right]
\end{equation}
$\psi(\cdot)$ is the probability of a gate being non-zero under the hard concrete distribution. It is more efficient in the second expectation to sample from the equivalent 
Bernoulli parameterization compared to hard concrete, which is more computationally expensive to sample multiple times. The first term now allows for efficient optimization 
via the reparameterization trick~\cite{kingma2013auto}; for the second, we apply the score function estimator (REINFORCE)~\cite{williams1992simple}, since the FLOPs
objective is, in general, non-differentiable and thus precludes the reparameterization trick. High variance is a non-issue because the number of
FLOPs is fast to compute, hence letting many samples to be drawn. At inference time, the deterministic estimator is
$\hat{\bftheta} := \bftheta \odot \max(0, \min(1, \text{Sigmoid}(\log \bfalpha) (\zeta - \gamma) + \gamma))$ for the final parameters $\hat{\bftheta}$.

{\bf FLOPs under group sparsity. }In practice, computational savings are achieved only if the model is sparse across ``regular'' groups of parameters, e.g., each filter in a 
convolutional layer. Thus, each computational group uses one hard concrete r.v.~\cite{louizos2018learning}---in fully-connected layers, one
per input neuron; in 2D convolution layers, one per output filter. Under convention in the literature where one addition and one multiplication
each count as a FLOP, the FLOPs for a 2D convolution layer $h_{conv}(\cdot;\bftheta)$ given a random draw $\bfz$ is then defined as 
$L_{flops}(h_{conv}, \bfz) = (K_w K_h C_{in} + 1) (I_w - K_w + P_w + 1) (I_h - K_h + P_h + 1) \lVert \bfz \rVert_0$
for kernel width and height $(K_w, K_h)$, input width and height $(I_w, I_h)$, padding width and height $(P_w, P_h)$, and number of input 
channels $C_{in}$. The number of FLOPs for a
fully-connected layer $h_{fc}(\cdot;\bftheta)$ is $L_{flops}(h_{fc}, \bfz) = (I_n + 1) \lVert \bfz \rVert_0$, where $I_n$ is the number of input neurons.
Note that these are conventional definitions in neural network compression papers---the objective can easily use instead a number of FLOPs incurred by 
other device-specific algorithms. Thus, at each training step, we compute the FLOPs
objective by sampling from the Bernoulli r.v.'s and using the aforementioned definitions, e.g., $L_{flops}(h_{conv}, \cdot)$ for convolution layers. Then, we apply the score 
function estimator to the FLOPs objective as a black-box estimator.

\section{Experimental Results}

We report results on MNIST, CIFAR-10, and CIFAR-100, training multiple models on each dataset corresponding to different FLOPs targets.
We follow the same initialization and hyperparameters as Louizos et al.~\cite{louizos2018learning}, using Adam~\cite{kingma2014adam} with temporal averaging for optimization, a weight
decay of $5 \times 10^{-4}$, and an initial $\alpha$ that corresponds to the original dropout rate of that layer. We similarly choose $\beta = 2/3$, 
$\gamma=-0.1$, and $\zeta=1.1$. For brevity, we direct the interested reader to their repository\footnote{\label{footnote:repo}\url{https://github.com/AMLab-Amsterdam/L0_regularization}} 
for specifics. In all of our experiments, we replace the original $L_0$ penalty with our FLOPs objective, and we train all models to 200 epochs; at 
epoch 190, we prune the network by weights associated with zeroed gates and replace the r.v.'s with their deterministic estimators, then finetune for 10 more epochs.
For the score function estimator, we draw 1000 samples at each optimization step---this procedure is fast and has no visible effect on training time.

\begin{table}[h]
    \caption{Comparison of LeNet-5-Caffe results on MNIST}
    \label{table:mnist}
    \centering
    \begin{tabular}{llcr}
        \toprule[1pt]
        Model & Architecture & Err. & FLOPs\\
        \midrule
        GL~\cite{wen2016learning} & 3-12-192-500 & 1.0\% & 205K\\
        GD~\cite{srinivas2016generalized} & 7-13-208-16 & 1.1\% & 254K\\
        SBP~\cite{neklyudov2017structured} & 3-18-284-283 & 0.9\% & 217K\\
        BC-GNJ~\cite{louizos2017bayesian} & 8-13-88-13 & 1.0\% & 290K\\
        BC-GHS~\cite{louizos2017bayesian} & 5-10-76-16 & 1.0\% & 158K\\
        $L_0$~\cite{louizos2018learning} & 20-25-45-462 & 0.9\% & 1.3M\\
        $L_0$-sep~\cite{louizos2018learning} & 9-18-65-25 & 1.0\% & 403K\\
        \midrule
        $L_{flops}$, $T=400$K & 3-13-208-500 & 0.9\% & 218K\\
        $L_{flops}$, $T=200$K & 3-8-128-499 & 1.0\% & 153K\\
        $L_{flops}$, $T=100$K & 2-7-112-478 & 1.1\% & 111K\\
        \bottomrule[1pt]
    \end{tabular}
\end{table}

We choose $\lambda_f = 10^{-6}$ in all of the experiments for LeNet-5-Caffe, the Caffe variant of LeNet-5.\footnotemark[1] We observe that our methods 
(Table \ref{table:mnist}, bottom three rows) achieve accuracy comparable to those from previous approaches while using fewer FLOPs, with the added benefit of 
providing a tunable ``knob'' for adjusting the FLOPs. Note that the convolution layers are the most aggressively compressed, since they 
are responsible for most of the FLOPs in this model.

\begin{table}[h]
    \caption{Comparison of WideResNet-28-10 results on CIFAR-10 and CIFAR-100}
    \centering
    \label{table:cifar}
    \begin{tabular}{lcccccc}
        \toprule[1pt]
        \multirow{2}{*}{\raisebox{-3\heavyrulewidth}{Method}} &
        \multicolumn{3}{c}{CIFAR-10} & \multicolumn{3}{c}{CIFAR-100}\\
        \cmidrule(lr){2-4}
        \cmidrule(lr){5-7}
         & Err. & $\EE$[FLOPs] & FLOPs & Err. & $\EE$[FLOPs] & FLOPs\\
        \midrule
        Orig. & 4.00\% & 5.9B & 5.9B & 21.18\% & 5.9B & 5.9B\\
        Orig. w/dropout & 3.89\% & 5.9B & 5.9B & 18.85\% & 5.9B & 5.9B\\
        $L_0$ & 3.83\% & 5.3B & 5.9B & \bfblue{18.75\%} & 5.3B & 5.9B\\
        $L_0$-small & 3.93\% & 5.2B & 5.9B & 19.04\% & 5.2B & 5.9B\\
        \midrule
        $L_{flops}$, $T=4$B & \bfblue{3.82\%} & 3.9B & 4.6B & 18.93\% & 3.9B & 4.6B\\
        $L_{flops}$, $T=2.5$B & 3.91\% & 2.4B & 2.4B & 19.48\% & 2.4B & 2.4B\\
        \bottomrule[1pt]
    \end{tabular}
\end{table}

Orig. in Table \ref{table:cifar} denotes the original WRN-28-10 model~\cite{zagoruyko2016wide}, and $L_0$-* refers to the $L_0$-regularized models~\cite{louizos2018learning}; 
likewise, we augment CIFAR-10 and CIFAR-100 with standard random cropping and horizontal flipping.
For each of our results (last two rows), we report the median error rate of five different runs, executing a total of 20 runs across two models for each
of the two datasets; we use $\lambda_f = 3 \times 10^{-9}$ in all of these experiments. We also report both the expected 
FLOPs and actual FLOPs, the former denoting the number of FLOPs, on average, at training time under stochastic gates and the latter denoting the number of FLOPs at inference 
time. We restrict the FLOPs calculations to the penalized non-residual convolution layers only. For CIFAR-10, our approaches result in Pareto-better models with decreases in both 
error rate and the actual number of inference-time FLOPs. For CIFAR-100, we do not achieve a Pareto-better model, since our approach trades accuracy for improved efficiency. The 
acceptability of the tradeoff depends on the end application.

\end{document}